\pdfoutput=1 
\documentclass[conference]{IEEEtran}
\IEEEoverridecommandlockouts

\usepackage{cite}
\usepackage{threeparttable}
\usepackage{amsmath,amssymb,amsfonts}
\usepackage{enumitem}
\usepackage{multirow}
\usepackage[super]{nth}
\usepackage[table,x11names]{xcolor}
\usepackage{float}
\usepackage{siunitx}
\usepackage{algorithmicx}
\usepackage{algpseudocode}
\usepackage{algorithm}
\usepackage{algpseudocode}
\usepackage{graphicx}
\usepackage{textcomp}
\usepackage{xcolor}
\usepackage{pifont}
\usepackage{makecell}
\usepackage{booktabs}
\usepackage{todonotes}
\usepackage[hidelinks]{hyperref}
\usepackage{balance}

\usepackage[T1]{fontenc}
\usepackage[a-1b]{pdfx}

\newcommand{\red}[1]{{\color{black}#1}} 
\newcommand{\blue}[1]{{\color{black}#1}}

\definecolor{softgreen}{RGB}{34, 139, 34} 

\begin{document}
\bstctlcite{IEEEexample:BSTcontrol} 
\title{

{\vspace{-0.6cm}\small This article is accepted for publication in \textit{ IEEE/ACM International Symposium on Low Power Electronics and Design} (ISLPED 2025) \\
}
\vspace{-0.8\baselineskip}
\rule{\textwidth}{0.4pt}\\
Exploration of Low-Power Flexible Stress Monitoring Classifiers for Conformal Wearables\vspace{-1ex}}

\author{
    \IEEEauthorblockN{
        Florentia Afentaki\IEEEauthorrefmark{1}, 
        Sri Sai Rakesh Nakkilla\IEEEauthorrefmark{2}, 
        Konstantinos Balaskas\IEEEauthorrefmark{1}, 
        Paula Carolina Lozano Duarte\IEEEauthorrefmark{3},\\ 
        Shiyi Jiang\IEEEauthorrefmark{4}, 
        Georgios Zervakis\IEEEauthorrefmark{1}, 
        Farshad Firouzi\IEEEauthorrefmark{2}, 
        Krishnendu Chakrabarty\IEEEauthorrefmark{2}, 
        Mehdi B. Tahoori\IEEEauthorrefmark{3}
    }
    
    \IEEEauthorblockA{
    \IEEEauthorrefmark{1}University of Patras, GR
    \IEEEauthorrefmark{2}Arizona State University, US
    \IEEEauthorrefmark{3}Karlsruhe Institute of Technology, DE
    \IEEEauthorrefmark{4}Duke University, US
    }

    \IEEEauthorblockA{
    \IEEEauthorrefmark{1}\{afentaki, kompalas, zervakis\}@ceid.upatras.gr, 
    \IEEEauthorrefmark{2}\{snakkill, Farshad.Firouzi, Krishnendu.Chakrabarty\}@asu.edu,    \\
    \IEEEauthorrefmark{2}\{paula.duarte, mehdi.tahoori\}@kit.edu
    \IEEEauthorrefmark{4}\{shiyi.jiang\}@duke.edu
    }
\vspace{-5ex}
}


\maketitle


\begin{abstract}
Conventional stress monitoring relies on episodic, symptom-focused interventions, missing the need for continuous, accessible, and cost-efficient solutions. 
State-of-the-art approaches use rigid, silicon-based wearables, which, though capable of multitasking, are not optimized for lightweight, flexible wear, limiting their practicality for continuous monitoring. 
In contrast, flexible electronics (FE) offer flexibility and low manufacturing costs, enabling real-time stress monitoring circuits.
However, implementing complex circuits like machine learning (ML) classifiers in FE is challenging due to integration and power constraints. 
Previous research has explored flexible biosensors and ADCs, but classifier design for stress detection remains underexplored.
This work presents the first comprehensive design space exploration of low-power, flexible stress classifiers. 
We cover various ML classifiers, feature selection, and neural simplification algorithms, with over \red{1200} flexible classifiers. 
To optimize hardware efficiency, fully customized circuits with low-precision arithmetic are designed in each case.
Our exploration provides insights into designing real-time stress classifiers that offer higher accuracy than current methods, while being low-cost, conformable, and ensuring low power and compact size.
\end{abstract}

\IEEEpubidadjcol



\section{Introduction}
\label{sec:introduction}
Stress is a critical health concern, linked to conditions such as depression, heart disease, digestive issues, and sleep disturbances~\cite{McEwen2017NeurobiologicalAS}.
Traditional stress monitoring methods, based on intermittent evaluations, fall short in providing the continuous data needed for accurate, timely analysis. 
Real-time monitoring, enabled by wearable devices processing physiological data from biosensors, is crucial for early intervention and better management of stress-related health issues. 
These devices often use sensory time-series data and machine learning (ML) algorithms to predict stress states.
However, most research has focused on algorithmic solutions~\cite{Kumar2021HierarchicalDN,Aqajari2020GSRAF, Lopez:IEEE:stressDataset:affectiveROAD, affectiveRoadComparison} or general-purpose microprocessor-based systems~\cite{Shiyi:IoT2022:StressMonitoring}, with limited attention to the hardware implications of wearable stress-monitoring devices. 
Traditional silicon-based wearable solutions, while effective, face limitations in terms of rigidity, high manufacturing costs, and substantial power consumption, making them unsuitable for continuous and accessible health monitoring applications.

Flexible electronics~(FE) offer a promising alternative for wearable  health monitoring devices, offering distinct advantages over conventional rigid electronics.
Their flexible substrates allow them to adapt more naturally to body contours, improving comfort during prolonged use~\cite{Gao:Nature2022:FlexibleSensor}.
Furthermore, they support the development of inexpensive, disposable hardware, making them well-suited for single-use low-cost patches in both commercial and medical applications.
However, FE exclusively use only n-type transistors in thin-film transistor~(TFT) technology, restricting circuit designs to unipolar logic.
Further, static power accounts for over $99\%$ of the total consumption of FE systems~\cite{ozer:nature2024:bendableRiscV}.
These limitations significantly hinder the integration capabilities of FE systems~\cite{ozer:nature2024:bendableRiscV}.

Significant research has been focused on mechanically-flexible biosensors capable of capturing stress-related signals, such as electrodermal activity~(EDA)~\cite{Kaipu:ICM2017:FlexibleEDA}.
However, the ML stress classifier--a critical component for stress prediction~\cite{Kumar2021HierarchicalDN}--remains largely unexplored by the state of the art.
This can be attributed to the inherent design constraints of FE in realizing such complex circuits.
\textit{Overall, matching the computational requirements of the healthcare applications, like stress prediction, with the unique mechanical characteristics of flexible technology remains an open question.
}

In this work, we address this research gap by thoroughly exploring the design space of ML classifiers for real-time stress monitoring systems.
Our aim is to design a set of ML classifiers that provide favorable accuracy-power tradeoffs, whilst complying with the stringent area constraints of FE.
To that end, we develop a custom standard cell library for FE, optimized for ultra-low-power operation at \SI{1}{\volt}, significantly lower than commercial solutions at \SI{3}{\volt}~\cite{PragmatIC:2023:NativelyFlexible}, enabling more power-efficient circuits.
We explore a wide design space comprising of
i) different ML algorithms, including Decision Trees~(DTs), Multilayer Perceptrons~(MLPs), and Support Vector Machines~(SVMs),
ii) statistical-based feature selection techniques,
and iii) neural minimization techniques such as pruning and low-precision quantization.
Leveraging the low cost of FE, we fully customize our circuits to each ML model, i.e., bespoke hardware implementation~\cite{Armeniakos:TC2023:codesign}, which greatly contributes to reducing power/area overheads.

\textbf{Our novel contributions within this work are as follows}:
\begin{enumerate}[topsep=0pt,leftmargin=*]
    \item To the best of our knowledge, this work is the first to design stress classifiers\footnote{Our stress classifiers are available at \url{https://github.com/floAfentaki/EDA-Driven-ML-Circuits-for-Flexible-Electronics}}
 in flexible electronics.
    \item We introduce an automated design space exploration (DSE) framework, evaluating over $1200$ classifiers, that integrates both software and hardware optimizations, including feature selection, neural minimization, bespoke circuit design, and low-precision arithmetic, in order to identify optimal power-accuracy trade-offs. 
    \item Our work highlights the feasibility of real-time flexible stress monitoring:
    our DSE identified solutions with \red{\SI{9}{\micro\watt}} power and \red{\SI{0.2}{\milli\meter}$^2$}  area, satisfying battery and area constraints in flexible electronics.
\end{enumerate}


\section{Flexible Electronics}  
\label{sec:background_flexible_electronics}  

The Indium Gallium Zinc Oxide (IGZO) Thin-Film Transistors (TFTs), incorporated into Flexible Integrated Circuit (FlexIC) technology, are advancing flexible electronics by combining mechanical adaptability with cost-efficient manufacturing~\cite{ozer:nature2024:bendableRiscV}.
Unlike conventional silicon-based solutions, IGZO TFTs can be fabricated on lightweight flexible substrates (e.g., polyimide) using low-temperature lithography, eliminating the need for additional protective packaging. 
This method not only avoids rigid silicon wafers and high-temperature fabrication but also significantly lowers production expenses and reduces environmental impact. 
Moreover, IGZO TFTs possess inherent mechanical flexibility, allowing them to bend without additional encapsulation. 
Their streamlined manufacturing process also shortens fabrication time from 32 weeks to under 3.5 days, making them suitable for scalable applications~\cite{pragmatic:whitepaper:sustainability}.  

Despite these advantages, IGZO TFTs face limitations compared to CMOS technology, particularly in terms of performance and feature size, with a typical minimum feature size of \SI{800}{\nano\meter}, significantly larger than that of silicon transistors~\cite{ozer:nature2024:bendableRiscV, pragmatic:ISSCC2022:Flex6502}.
Thus, designing complex circuits, such as ML classifiers, for applications with strict area constraints, like wearables, poses a significant challenge in FE.
Additionally, IGZO TFT technology relies solely on n-type transistors, restricting designs to unipolar logic. 
Specifically, resistor-NMOS (R-NMOS) logic is utilized, where a pull-up resistor replaces the PMOS transistor. 
The absence of p-type devices increases resistance, affecting delay and power consumption, leading to design challenges.
To mitigate these, hardware-software co-design strategies, such as fully parallel bespoke (see Section \ref{sec:bespoke}) implementations and power-efficient logic simplifications, are necessary~\cite{Ozer:2019:Bespoke}.
Our approach incorporates these strategies to optimize flexible components, achieving reductions in memory usage, gate count and power consumption, eliminating also the need for memory elements that are scarce and coslty in FE technology—key factors for developing efficient, lightweight healthcare monitoring systems.
\section{Designing A Stress Monitoring System}

\subsection{System Overview}
\label{sec:system_overview}
\begin{figure}[!t]
\centering
\includegraphics[width=0.8\columnwidth]{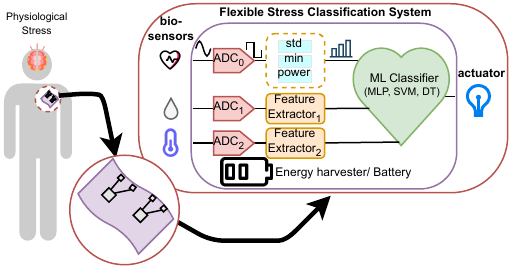}
\vspace{-2ex}
\caption{Overview of the mechanically flexible real-time stress monitoring classification system.}
\label{fig:system_overview}
\vspace{-3ex}
\end{figure}

\figurename~\ref{fig:system_overview} presents an abstract block diagram of our targeted flexible classification system for real-time stress monitoring. 
Flexible biosensors~\cite{Moy:SSC2017:EEGFlexible} capture bio-signals, which are quantized by Analog-to-Digital Converters~(ADCs)~\cite{duarte:2024:prunedADC} and processed by feature extractors to generate input features for the classifier.
The trained ML classifier processes these features and predicts stress levels. 
\textit{To the best of our knowledge, this is the first work to systematically design and optimize classifiers specifically for stress prediction in FE.}

Significant research on FE has focused on flexible sensors~\cite{Kaipu:ICM2017:FlexibleEDA} and ADCs~\cite{duarte:2024:prunedADC}, while the feature extractor could be implemented with a flexible microprocessor like in~\cite{ozer:nature2024:bendableRiscV}.
However, a thorough investigation of the impact of feature selection (i.e., number of features and selection method) on the flexible ML classifier is missing from the literature.
Additionally, \textit{no prior work has systematically explored neural simplification techniques (e.g., pruning and quantization) alongside classifier selection to minimize the hardware footprint of a flexible classifier.  }
In our work, we focus on optimizing the classifier within the context of the stress monitoring system.

\subsection{Bespoke Flexible Classifier Design}\label{sec:bespoke}
Targeting to comply with strict area constraints (i.e., a primary design objective of FE), we design fully-parallel bespoke ML circuits.
Bespoke refers to hardwiring the model coefficients to the hardware implementation, significantly boosting the efficiency compared to conventional designs~\cite{Mubarik:MICRO:2020:printedml}.
Such customization is enabled by the low non-recurring engineering~(NRE) and fabrication costs of FE.
Additionally, fully-parallel designs are purely combinational, alleviating the need for excess memory units, which are costly in FE~\cite{ozer:nature2024:bendableRiscV}.

\label{sec:designing}

Based on these guidelines, each ML algorithm demands its own tailored implementation.
MLPs compute a weighted sum per neuron, adding a final bias.
Our fully-parallel MLPs instantiate one bespoke multiplier per weight and a semi-bespoke adder-tree (i.e., accumulation of bespoke products), where the weight and bias are known a priori and set as constants, leading to much simplified circuitry.
SVMs with a linear kernel operate on the same principle, computing a weighted sum; however, in this case, a separate weighted sum is calculated for each output class i.e., binary classifier.
Finally, DT circuits comprise a series of comparators, where input features are compared against hardwired thresholds in parallel, determining the activated tree branches and finally, the predicted class.
\textit{This work demonstrates the feasibility of fully-parallel and mechanically-flexible classifiers targeting real-time stress prediction.}


\section{Proposed Flexible Classifier Exploration}

\begin{figure}[!t]
\centering
\includegraphics[width=\columnwidth]{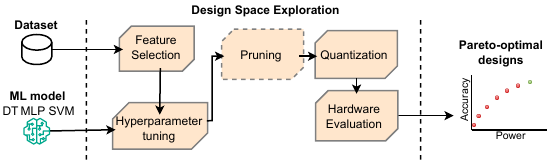}
\vspace{-4ex}
\caption{Algorithmic flowchart of our proposed design space exploration.}
\vspace{-3ex}
\label{fig:dse}
\end{figure}

We propose a DSE which aims to identify accuracy-power Pareto-optimal designs for real-time stress monitoring classifiers, focusing on design feasibility under the constraints of FE.
We present the pseudocode for our flexible classifier design space exploration in Algorithm~\ref{alg:dse}, while the algorithmic flowchart is shown in \figurename~\ref{fig:dse}. 
\begin{algorithm}[t!]
\caption{Flexible Classifier Design Space Exploration}
\label{alg:dse}
\footnotesize
\begin{algorithmic}[1]
\Require Feature set $X$, labels $Y$
\Ensure Optimized classifier $C^*$
\State Initialize feature set $S = X$
\For{each feature selection $F \in \{\text{DISR, Fisher, JMI}\}$}
    \State Select top-$k$ features $S_F = F(X, Y)$
    \For{each ML classifier $C \in \{\text{SVM, MLP, DT}\}$}
        \For{each hyperparameter set $H_C$}
            \State Train classifier $C$ using $S_F$ and $H_C$
            \If{$C$ is an MLP}
                \For{each sparsity $s \in \{0.2, 0.5, 0.9\}$}
                    \State Apply pruning-aware retraining with $s$
                \EndFor
            \EndIf
            \For{each precision $p \in \{4,6,8,10\}$}
                \State Apply quantization with $p$-bit precision
                \State Obtain Hardware Description
                \State Apply Hardware Evaluation
                \State Obtain Accuracy $A_C$
                \State Obtain Power consumption $P_C$
            \EndFor
        \EndFor
    \EndFor
\EndFor
\State Return $C^* = Pareto_{C} ( A_C, P_C )$
\end{algorithmic}
\end{algorithm}
Our DSE incorporates techniques such as feature selection, pruning (for MLPs), and low-precision arithmetic, to optimize hardware efficiency and enable classifiers' realization in FE implementations.
For each ML algorithm, we first apply statistical feature selection, using state-of-the-art techniques, to identify a subset of the most relevant statistics, and then train the respective classifier.
Quantization is used to explore low-precision implementations across all classifiers, reducing the hardware overheads, while state-of-the-art unstructured pruning is also applied to the MLPs.
The above form a complex design space of software-hardware design techniques, all aiming to reduce the footprint of the flexible classifier without deteriorating the application accuracy.
Finally, the hardware description of the stress classifiers is obtained via custom Python-to-Verilog code templates in a fully automated way, and their hardware evaluation is conducted with a low-power standard cell library via commercial EDA tools.
\textit{To the best of our knowledge, this is the first time such an exploration is conducted for stress monitoring within FE.
}

\subsection{Feature Selection}
\label{sec:feature_selection}
Feature selection significantly impacts both hardware efficiency and classification accuracy.
Limiting the number of selected features directly reduces the classifier's inputs and consequently its size and parameters, lowering the associated overheads.
Additionally, choosing a technique that identifies and retains only the most relevant features can allow for higher achievable accuracy.
Accounting for the above, we simultaneously explore the feature selection algorithm and the number of selected features, in order to achieve optimal accuracy while satisfying the strict constraints of FE.

\label{sec:exploration}

Our feature selection is conducted offline, and three state-of-the-art statistical-based algorithms are explored~\cite{Shiyi:IoT2022:StressMonitoring}:
Double Input Symmetrical Relevance~(DISR), Fisher Score, and Joint Mutual Information~(JMI).
The Fisher Score evaluates the discriminative power of features by comparing inter-class to intra-class variance, where, higher scores indicate stronger class separation.
JMI evaluates the shared information between the features, while DISR normalizes JMI by the combined entropy of the features, emphasizing those with unique information.
Taking into account the uncertainty in the system, DISR highlights features that contribute distinct information.
Given a dataset with $M$ features $X = \{x_1, x_2, \dots, x_M\}$ and corresponding labels $Y$, we define an optimization objective to select a subset $S^*$ of $k$ features that maximizes a relevance criterion $F(S)$:
\begin{equation}
    S^* = \arg\max_{S \subset X, |S| = k} F(S).
\end{equation}
where \(F(S)\) represents the relevance criterion computed using one of the feature selection algorithms (DISR, Fisher Score, or JMI). This process ensures that only the most important features are retained, maintaining high accuracy while reducing hardware requirements.
\blue{Beyond the algorithm exploration, we evaluate different numbers of features per algorithm by varying $k$ within a predefined limit, ensuring that our flexible classifiers' input size stays within reasonable area bounds.}


\subsection{Classifier Training and Neural Minimization Techniques}
\label{sec:training}
Following the feature selection process, the classifiers are then trained with the selected features.
During training, \emph{hyperparameter tuning} is performed to identify the optimal model architecture for each considered ML algorithm, and obtain the most accurate model.
SVMs are restricted to use only linear kernels, avoiding complex non-linear alternatives which induce significant hardware overheads.
MLPs use the Rectified Linear Unit~(ReLu) activation function due to its simplicity and low hardware requirements.
Finally, different splitting criteria for DTs are explored, including gini, impurity, and entropy.

\emph{Quantization} is then applied on the trained classifier to mimic the effect of precision scaling in hardware, leveraging the efficiency of low-precision arithmetic.
As precision decreases, the hardware overhead of all arithmetic components (i.e., multipliers/adders in MLPs and SVMs, comparators in DTs) decreases accordingly at the cost of accuracy loss.
Importantly, our bespoke designs enable the precision of coefficients and input features to be tailored to specific application requirements, allowing for a maximum exploitation of low-precision arithmetic while adhering to high accuracy constraints.

As an additional minimization technique, we consider \emph{pruning-aware retraining} for MLPs, due to their elevated hardware cost.
Unstructured pruning removes unimportant coefficients during training, leading to smaller models whilst preserving accuracy.
In hardware, the multipliers corresponding to pruned coefficients are removed from our bespoke fully-parallel circuits~\cite{Armeniakos:TC2023:codesign}, while the adder tree is simplified by accumulating less addends.
For example, if we consider a bespoke neuron with 5 inputs, the weighted sum would be \(in_0c_0 + in_1c_1 + in_2c_2 + in_3c_3 + in_4c_4.\)
After pruning, if some weights are set to zero, e.g., $c_1 = 0$ and $c_4 = 0$, the neuron's output becomes \(in_0c_0 + in_2c_2 + in_3c_3\), saving two multipliers and two operands of the adder tree.
We evaluate three state-of-the-art pruning techniques, each employing a distinct ranking criterion: L2-norm, Hessian, and activation-aware pruning. 
In L2 Norm method, weights with smaller absolute magnitudes are pruned, as smaller weights contribute less to each neuron's output. 
Hessian-Based pruning removes weights with smaller second-order derivatives, represented by the diagonal elements of the Hessian matrix, as these weights have minimal impact on reducing the model's loss function. 
Activation-Aware considers both the magnitude of weights and the norms of their corresponding activations, pruning based on their combined importance.
An exhaustive exploration as a calibration step demonstrates that pruning-aware retraining using the L2-norm criterion consistently outperforms the other two techniques.
Thus, the L2-norm criterion is selected for further exploration. 
Finally, we explore different sparsity ratios to cover a wide range of possible accuracy-power tradeoffs.
\blue{Given the small sizes of ML models designed for FE and the high-level nature of the optimizations, the design space is exhaustively explored in parallel, enabling fast evaluation and generalization to other ML applications of FE.}

\section{Results \& Analysis} \label{sec:evaluation}


\subsection{Flexible Standard Cell Library Characterization}
\label{sec:library}
\begin{figure}[t!]
\centering
\includegraphics[width=0.9\columnwidth]{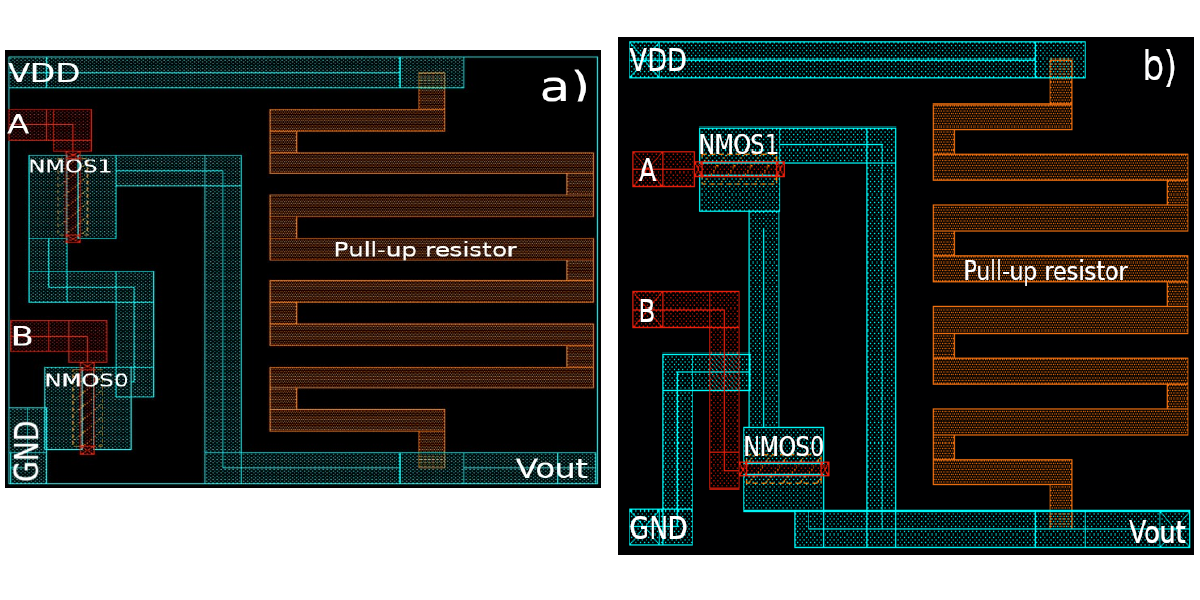}
\vspace{-0.6cm}
\caption{Resistor-NMOS layout of 2-input a) NAND and b) NOR gates.}
\label{fig:layout}
\vspace{-3ex}
\end{figure}

In this work, we develop and characterize a custom standard cell library for FE, optimized for ultra-low-power operation at \SI{1}{V} using the PragmatIC FlexICs PDK second-generation Helvellyn 2.1.0\cite{Europractice:FlexICs}. 
The target operating voltage of \SI{1}{\volt} was selected to minimize cell size and power consumption while achieving a \SI{1}{\micro\second} delay across all cells.
To address the inherent challenges of IGZO TFTs, careful adjustments to the transistor threshold voltages and drive strengths are implemented to guarantee proper switching behavior and functionality under low-voltage conditions.
Design considerations include reducing leakage currents, compensating for increased pull-up resistance, optimizing resistor placement for power efficiency and speed, and minimizing parasitic capacitances through interconnect optimization. 
\figurename~\ref{fig:layout} shows the layout of our R-NMOS 2-input NAND (left) and NOR (right) gates optimized for \SI{1}{V} operation. 
The area of all cells, obtained from final layouts, is reported in Table~\ref{tab:cell_areas}. 
Cells are characterized through simulations considering input slew and output load capacitance using the PDK's typical transistor model. 
Transition times, static and dynamic power, and drive strength are measured to ensure accurate modeling. 
The extracted data is compiled into a Liberty (\texttt{.lib}) file for integration with commercial EDA tools for delay, area, and power analysis.


\begin{table}[t!]
    \centering
    \caption{Area measurements of our primitive gates using the PragmatIC FlexICs PDK second generation Helvellyn 2.1.0\cite{Europractice:FlexICs}.}
    \setlength{\tabcolsep}{20pt} 
\renewcommand{\arraystretch}{1.0} 
    \begin{tabular}{|c|S[table-format=5.2]|}
        \toprule
        \textbf{Cell Name} & \textbf{Area (\si{\square\micro\meter})} \\
        \midrule
        INVX1 & 747.82 \\
        AND2  & 2121.00 \\
        NAND2 & 919.01 \\
        OR2 & 2468.85 \\
        NOR2  & 1053.62 \\
        DFFNRX1 & 16195.00 \\
        \bottomrule
    \end{tabular}
    \label{tab:cell_areas}
\vspace{-2ex}
    
\end{table}


\subsection{Stress Datasets}
\subsubsection{Wearable Stress and Affect Detection Dataset}
We evaluate our proposed flexible classifiers using the Wearable Stress and Affect Detection~(WESAD) dataset~\cite{Schmidt2018IntroducingWA}.
The collected data originate from two wearable devices on the chest (RespiBAN) and at the wrist (Empatica E4), worn by 17 participants for around 100 minutes.
In our exploration, all the signals from RespiBAN are used,
while from the Empatica E4 only BVP is used.
Each recorded sample was labeled in one of the two individual's stress level: baseline/rest and stress, i.e., binary classification.

\subsubsection{AffectiveROAD}

The AffectiveROAD dataset~\cite{MIT:stressDataset:affectiveROAD} focuses on stress monitoring in driving scenarios, integrating both physiological and contextual data collected from two devices: the Empatica E4 and Zephyr Bioharness 3. 
It includes data from 13 real-world driving sessions conducted by nine experienced drivers, self-identified as four women and five men.
For our analysis, only raw sensor data from the Empatica E4 is used.
\emph{These datasets enable us to validate our system across diverse real-world stress applications.}

\subsection{Experimental Setup}

Feature Extraction was performed using the Scipy and pyHRV packages~\cite{gomes:2019:pyhrv}.
For training, we normalize the extracted features within $[0,1]$ and then randomly split the formed dataset into training and testing subsets, with a $70\%/30\%$ split ratio. 
Stratification ensured a balanced distribution of each target class within both the training and testing sets.
Scikit-learn's GridSearchCV is used for hyperparameter selection during training with $5$-fold cross validation, and the ML models are trained until convergence with default tolerance.
For synthesis and mapping, we use the standard-cell library developed at $1\si{\volt}$ with the Pragmatic FlexIC PDK~\cite{duarte:2024:prunedADC} developed as mentioned in Section\ref{sec:library}.
Synopsys Design Compiler S-2021.06, VCS T-2022.06, and PrimeTime T-2022.03 are used for synthesis and hardware analysis.
The accuracy is reported on the test dataset, and all designs are synthesized at clock period of $2\si{\kilo\hertz}$, aligning with the performance of typical stress monitoring applications~\cite{wang2017flexible}.
\emph{Over \red{$1200$} classifiers are evaluated in our exploration.}

\subsection{Evaluation of our Flexible Classifiers}
\label{sec:classifier_evaluation}

\begin{figure}[t!]
\centering
\includegraphics[width=\columnwidth]{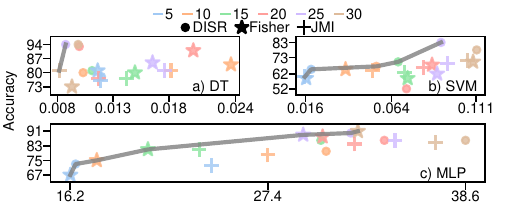}
\includegraphics[width=\columnwidth]{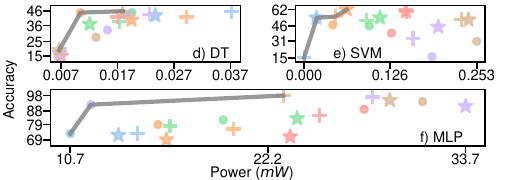}
\vspace{-4ex}
\caption{Feature selection evaluation of accuracy-power trade-offs, across all considered ML algorithms: (a, d) DTs, (b, e) SVMs, and (c, f) MLPs. The figures (a-c) presents the WESAD evaluation, while (d-f) studies AffectiveROAD dataset.
The coefficients of the ML models are 8-bit fixed-point values.
}
\label{fig:fs}
\vspace{-3ex}
\end{figure}


\subsubsection{Feature Selection Evaluation}
\label{sec:fs_evaluation}
First, we evaluate the impact of different feature selection techniques and number of selected features in designing low-area but highly accurate stress classifiers.
\figurename~\ref{fig:fs} 
presents the evaluation of feature selection for all three classifiers (DT, SVM and MLP), with 8-bit fixed-point coefficients, which has demonstrated optimal gains without accuracy loss in the state of the art~\cite{Afentaki:ICCAD23:hollistic}.
We explore a number of selected features within the range of $5$ to $30$, using increments of $5$.
The top of \figurename~\ref{fig:fs} illustrates the accuracy-area tradeoffs, while the bottom depicts the accuracy-power tradeoffs. 
\red{\figurename~\ref{fig:fs} (a-c) and (d-f) illustrates the accuracy-power tradeoffs of WESAD and AffectiveROAD dataset respectively.
}
Due to space limitations, the accuracy-area tradeoff figure is not included; however, the accuracy-power and accuracy-area tradeoffs would appear identical. In flexible electronics (FE), nearly $99\%$ of power consumption is static \cite{ozer:nature2024:bendableRiscV}, making area and power linearly correlated. 
Hence, area and power are linearly correlated and minimizing area also minimizes power since, unlike in conventional IC, the contribution of switching activity to the overall power consumption is negligible.

Overall, we observe that each Pareto front is populated by diverse feature selection techniques.
For SVMs and MLPs for the WESAD, $67\%$ and $71\%$ of Pareto-optimal points use DISR and Fisher, respectively, whereas Pareto solutions for DTs are divided into DISR and JMI.
\red{
For AffectiveROAD, $67\%$ of Pareto-optimal points for MLP use DISR, while Pareto solutions for MLPs are split between DISR and JMI, and for DT, the techniques are equally selected.
}
This indicates that the choice of selection method is not trivial and depends on the type of classifier, showcasing the necessity to explore different techniques per ML algorithm. 
Interestingly, we observe that the number of features does not necessarily correlate with the classifier's power, as fewer features may result in larger circuits.
This might happen since in bespoke circuits, where coefficient values define the area of instantiated arithmetic components, hardware overheads are highly influenced by both the number of trained parameters and their specific value.
Similar observations can be extended to accuracy, where adding more features does not necessarily enhance it, and the significance of different feature combinations varies w.r.t. the achieved accuracy.

\begin{figure}[t!]
\centering
\includegraphics[width=\columnwidth]{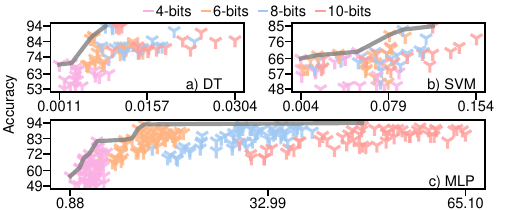}
\includegraphics[width=\columnwidth]{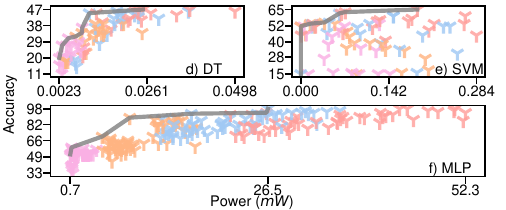}
\vspace{-4ex}
\caption{Accuracy-power evaluation of our DSE for generating flexible stress classifiers, using feature selection and neural minimization techniques, across all considered ML algorithms: (a, d) DTs, (b, e) SVMs, and (c, f) MLPs. The figure (a-c) presents the evaluation of WESAD, while (d-f) of AffectiveROAD.}
\label{fig:results}
\vspace{-4ex}
\end{figure}

\subsubsection{DSE Evaluation}
\figurename~\ref{fig:results} presents the accuracy-power trade-offs obtained from our entire DSE, using different neural minimization approaches on top of feature selection, across all ML algorithms.
Specifically, quantization is explored for $4$, $6$, $8$, and $10$ bits, and L2-norm pruning with sparsity ratios of $20\%$, $50\%$, and $90\%$~(for MLPs only).
We observe that our flexible classifiers can highly benefit from quantization, as all accuracy-power Pareto fronts are populated by low-precision designs.
Specifically for WESAD, $43\%$ of all Pareto-optimal classifiers feature $4$-bit precision, $39\%$ use $6$ bits, and only $18\%$ require $8$ bits or more.
\red{For AffectiveROAD, $30\%$ of all Pareto-optimal classifiers feature $4$-bit precision, $26\%$ use $6$ bits, $26\%$ require $8$ bits and only $17\%$ $10$ bits.
}
This highlights the effectiveness of exploring low-precision arithmetic in our bespoke fully-parallel designs, as quantization highly influences which solutions become Pareto-optimal.
\red{We also observe that MLPs demonstrate better robustness to quantization, as they exhibit only a small accuracy drop at $6$ bits compared to $10$ bits.
Contrarily, reducing precision incurs significant drops for SVMs and DTs, of $2\%$ on average. }
\blue{This trend is illustrated in \figurename~\ref{fig:results}, where the MLP solutions exhibit greater robustness to reduced precision, with accuracy degrading more gracefully compared to the steeper drops observed for SVM and DT.}
Finally, pruning facilitates the removal of parameters, and therefore yields power gains in MLP circuits. 
The Pareto-optimal points for WESAD are distributed as follows: $45.5\%$ of the points correspond to a sparsity of $20\%$, $18.2\%$ to a sparsity of $50\%$, and $36.4\%$ to a sparsity of $90\%$.  
While \red{for the AffectiveROAD are distributed as follows: $29\%$ of the points correspond to a sparsity of $20\%$, $57.0\%$ to a sparsity of $50\%$, and $14\%$ to a sparsity of $90\%$.  } 
\red{Interestingly, the majority of Pareto-optimal points use $20\%$ and $50\%$ sparsity ratio, for WESAD and AffectiveROAD respectively, highlighting the fact that determining the best pruning ratio is not trivial.
}
It is important to reiterate that in bespoke architectures, area and power overheads depend on the coefficient values. 
During pruning-aware retraining, accuracy recovery may favor less hardware-friendly coefficients, leading to less pruning and fewer changes, better suited for bespoke architectures.

\begin{table}[t!]
\centering
\caption{Comparison of our Highly-Accurate Flexible Classifiers}
\setlength{\tabcolsep}{4pt} 
\renewcommand{\arraystretch}{1.1} 
\begin{threeparttable}
\begin{tabular}{|l|ccc|ccc|}
    \toprule
        \textbf{Dataset} & \multicolumn{3}{c}{\textbf{WESAD}} & \multicolumn{3}{c|}{\textbf{AffectiveRoad}}\\
    \textbf{Model} & \textbf{MLP$^1$} & \textbf{SVM} & \textbf{DT} & \textbf{MLP$^2$} & \textbf{SVM} & \textbf{DT}\\
    \midrule
\textbf{Feature Selection}
    & Fisher & DISR & DISR & JMI & Fisher & DISR\\
    \textbf{\#Features} & 25 & 25 & 25 & 30 & 20 &15 \\
    \textbf{Precision} & 10 & 10 & 8 & 8 & 10 & 10\\
   \textbf{Accuracy (\%)}& 94 & 85 & 94  & 98  & 65 & 47\\
   \textbf{F1 Score (\%)}& 93 & 85 & 94 & 100 & 86 & 99 \\
\textbf{Area (\si{\square\centi\meter})} & 8.8 & 0.021 & 0.002 & 9.3 & 0.065 & 0.009\\
    \textbf{Power} (\si{\milli\watt}) & 48 & 0.12 & 0.009 &  26.5  &0.19 & 0.025\\
    \textbf{Latency} (\si{\milli s}) & 6.3 & 0.7 & 0.14 & 97 & 15 & 7.1\\
    \textbf{Energy} (\si{\micro\joule}) & 300 & 0.08 & 0.001 & 2600 & 2.85 & 0.1775
    \\
    \bottomrule
\end{tabular}
\begin{tablenotes}
   \item[]$^1$MLP is pruned with L2-norm criterion and $90\%$ sparsity. $^2$MLP is pruned with L2-norm criterion and $50\%$ sparsity.
\end{tablenotes}
\end{threeparttable}
\label{tab:accuracy_comparison}
\vspace{-4ex}
\end{table}

\subsubsection{Evaluation of our Most Accurate Classifiers}
Prioritizing application accuracy, Table~\ref{tab:accuracy_comparison} provides a detailed analysis of our flexible classifiers that achieve the highest accuracy for each ML algorithm, as obtained by our DSE.
For WESAD, we observe that \red{MLPs} and \red{DTs} achieve the highest accuracy.
However, DTs are also the most hardware-efficient, requiring on average only \red{\SI{0.2}{\square\milli\meter}} and consuming just \red{\SI{9}{\micro\watt}} with only \red{\SI{0.001}{\micro\joule} energy per inference.}
Both DTs and SVMs consume less than \SI{2}{\milli\watt} of power, allowing them to be powered by existing flexible energy harvesters. 
This enables self-sustaining operation, a key advantage for wearable applications.
Our most accurate MLP, even after pruning $90\%$ of its parameters, still incurs high hardware costs, 
with power consumption exceeding \SI{30}{\milli\watt} i.e, non-adequate for battery-powered operation.
It is important to underscore that, while energy consumption is often a consideration, power availability is the more critical constraint in FE. 
Given that printed batteries can be customized in capacity, shape, and voltage~\cite{PrintedBatteries2018}, managing peak power consumption is a higher priority than optimizing for total energy usage~\cite{Henkel:ICCAD2022:expedition}.
\red{For AffectiveROAD, 
Both DTs and SVMs consume less than 0.1 \si{\milli\watt}, but their low accuracy limits their practical use. 
However, it is worth mentioning that SVM accuracy is comparable to other state-of-the-art approaches \cite{Lopez:IEEE:stressDataset:affectiveROAD}, while DTs have not been used in the literature.
In contrast, MLPs, despite higher hardware costs,
provide realistic accuracy while being able to operate with the largest printed battery i.e, Molex.}
\textit{The above highlights the importance of a design space exploration like ours, as different stress datasets require different ML models and design decisions, to balance accuracy and hardware constraints in FE.
}

\subsubsection{State-of-the-Art Comparison}
As mentioned in Section~\ref{sec:introduction}, the state of the art for stress monitoring uses rigid, silicon-based wearables focused on algorithmic design~\cite{Shiyi:IoT2022:StressMonitoring,Kumar2021HierarchicalDN,Aqajari2020GSRAF, Lopez:IEEE:stressDataset:affectiveROAD, affectiveRoadComparison}. 
These systems rely on general-purpose hardware (e.g., CPUs~\cite{affectiveRoadComparison}) and require continuous transmission of data to the cloud~\cite{Shiyi:IoT2022:StressMonitoring}. 
Table~\ref{tab:sota_comp} compares our flexible solutions with the rigid systems in~\cite{Shiyi:IoT2022:StressMonitoring, affectiveRoadComparison}. 
Our solution enables a fully-flexible classification system that can be realized as a standalone on-body patch in flexible technology. In contrast to rigid silicon, our system offers lower cost, significantly reduced power consumption, and purely edge-based computations.



The rigid silicon system in \cite{Shiyi:IoT2022:StressMonitoring} achieved $87\%$ accuracy for WESAD, using SVM as the optimal classifier for edge computing. 
However, it did not explore DTs or MLPs, which our work identifies as optimal (see also Table~\ref{tab:accuracy_comparison}). 
For the AffectiveROAD dataset, the authors in \cite{affectiveRoadComparison} achieved $81\%$ accuracy, focusing on algorithmic optimization and general-purpose CPUs. 
In contrast, we achieve state-of-the-art accuracy ($94\%$, $81\%$), surpassing previous silicon-based solutions~\cite{Shiyi:IoT2022:StressMonitoring, Aqajari2020GSRAF, Kumar2021HierarchicalDN, affectiveRoadComparison, Lopez:IEEE:stressDataset:affectiveROAD}. 
Our flexible classifiers, with minimal area overheads and battery-powered operation, can be seamlessly integrated into conformal and accessible wearable devices.




\begin{table}[t!]
\centering
\caption{State-of-the-Art Comparison of Stress Monitoring Classifiers}
\label{tab:sota_comp}
\renewcommand{\arraystretch}{1.1} 
\begin{tabular}{|l|c|c|}
\toprule
\textbf{Feature} & \textbf{Silicon-based~\cite{Shiyi:IoT2022:StressMonitoring, affectiveRoadComparison}} & \textbf{Our} \\ 
\midrule
\multirow{2}{*}{\textbf{Flexibility}} & \textcolor{red}{\ding{55} Rigid} & \textcolor{softgreen}{\ding{51} Fully flexible} \\  
 & \textcolor{red}{\ding{55} Bulky} & \textcolor{softgreen}{\ding{51} Patch-based} \\ \hline
\textbf{Cost} & \textcolor{red}{ $>10$ dollar} & \textcolor{softgreen}{ sub-dollar} \\ \hline
\textbf{Computational Model} & \textcolor{red}{ Cloud-dependent} & \textcolor{softgreen}{ Fully edge-based} \\ \hline
\textbf{Accuracy} & \textcolor{red}{ 87\%, 81\%} & \textcolor{softgreen}{ 94\%, 98\%} \\ 
\bottomrule
\end{tabular}
\vspace{-4ex}
\end{table}

\section{Conclusion}
\label{sec:conclusion}
In this work, we conduct the first comprehensive design space exploration of mechanically-flexible low-power classifiers for real-time stress monitoring applications.
To that end, we incorporate diverse ML algorithms in our exploration, accounting for the hardware impact of varied sets of features and neural simplification techniques, such as unstructured pruning and low-precision quantization.
Our flexible classifiers are designed as bespoke fully-parallel circuits, aiming to comply with the stringent constraints of FE.
We designed and evaluated over $1200$ classifiers within our exploration.
Our results reveal that our Pareto-optimal flexible classifiers enable personalized stress classification, achieving state-of-the-art accuracy with a small, conformal, and accessible device compared to rigid, state-of-the-art solutions.


\section*{Acknowledgment}
\small{
\blue{
This work is partially supported by the European Research Council (ERC) and co-funded by the H.F.R.I call “Basic Research Financing (Horizontal support of all Sciences)” under the National Recovery and Resilience Plan “Greece 2.0” (H.F.R.I. Project Number: 17048).
}
}

\balance

\bibliographystyle{IEEEtran}
\bibliography{IEEEabrv,references}

\end{document}